\documentclass[sigconf]{acmart}
\AtBeginDocument{%
  }

\copyrightyear{2026}
\acmYear{2026}
\setcopyright{cc}
\setcctype{by}
\acmConference[WWW '26]{Proceedings of the ACM Web Conference 2026}{April 13--17, 2026}{Dubai, United Arab Emirates}
\acmBooktitle{Proceedings of the ACM Web Conference 2026 (WWW '26), April 13--17, 2026, Dubai, United Arab Emirates}
\acmPrice{}

\usepackage{hyperref}
\usepackage{booktabs}
\usepackage{multirow}
\usepackage{array}
\usepackage{amsmath}
\usepackage{subcaption} 
\usepackage{algorithm}

\usepackage{newfloat}
\usepackage{listings}
\usepackage{fancyhdr}
\usepackage{bibentry}
\usepackage{pifont}
\usepackage{url}
\usepackage{algorithm}
\usepackage{algpseudocode}
\begin{document}

\title[Causal Neuro-Symbolic Reasoning Model for Explainable Multi-Behavior Recommendation]{Modeling Endogenous Logic: Causal Neuro-Symbolic Reasoning Model for Explainable Multi-Behavior Recommendation}



\author{Yuzhe Chen}
\orcid{0000-0001-8884-7780}
\affiliation{%
  \department{School of Computer Science and Engineering}
  \institution{Nanjing University of Science and Technology}
  \city{Nanjing}
  \country{China}
}
\email{chenyuzhe@njust.edu.cn}

\author{Jie Cao}
\authornote{Corresponding author.}
\orcid{0000-0002-7049-5614}
\affiliation{%
  \department{School of Management}
  \institution{Hefei University of Technology}
  \city{Hefei}
  \country{China}
}
\email{cao_jie@hfut.edu.cn}
\thanks{This work was supported by the National Natural Science Foundation of China (Grant Nos. 72342011 and 72172057).}
\author{Youquan Wang}
\orcid{0000-0003-4726-7493}
\affiliation{%
  \department{School of Computer Science and Artificial Intelligence}
  \institution{Nanjing University of Finance and Economics}
  \city{Nanjing}
  \country{China}
}
\email{youq.wang@gmail.com}

\author{Haicheng Tao}
\orcid{0000-0002-1286-2578}
\affiliation{%
  \department{School of Computer Science and Artificial Intelligence}
  \institution{Nanjing University of Finance and Economics}
  \city{Nanjing}
  \country{China}
}
\email{haicheng.tao@nufe.edu.cn}

\author{Darko B. Vukovic}
\orcid{0000-0002-1165-489X}
\affiliation{%
  \department{Department of Finance and Accounting}
  \institution{Saint Petersburg State University}
  \city{St. Petersburg}
  \country{Russian Federation}
}
\email{d.vukovic@gsom.spbu.ru}

\author{Jia Wu}
\orcid{0000-0002-1371-5801}
\affiliation{%
  \department{School of Computing}
  \institution{Macquarie University}
  \city{Sydney}
  \country{Australia}
}
\email{jia.wu@mq.edu.au}

\renewcommand{\shortauthors}{Y. Chen et al.}

\renewcommand{\shortauthors}{Yuzhe Chen et al.}

\begin{abstract}
Existing multi-behavior recommendations tend to prioritize performance at the expense of explainability, while current explainable methods suffer from limited generalizability due to their reliance on external information. Neuro-Symbolic integration offers a promising avenue for explainability by combining neural networks with symbolic logic rule reasoning. Concurrently, we posit that user behavior chains (e.g., view$\to$cart$\to$buy) inherently embody an endogenous logic suitable for explicit reasoning. However, these observational multiple behaviors are plagued by confounders, causing models to learn spurious correlations. By incorporating causal inference into this Neuro-Symbolic framework, we propose a novel \textbf{C}ausal \textbf{N}euro-Symbolic \textbf{R}easoning model for \textbf{E}xplainable Multi-Behavior Recommendation (CNRE). CNRE operationalizes the endogenous logic by simulating a human-like decision-making process. Specifically, CNRE first employs hierarchical preference propagation to capture heterogeneous cross-behavior dependencies. Subsequently, it models the endogenous logic rule implicit in the user's behavior chain based on preference strength, and adaptively dispatches to the corresponding neural-logic reasoning path (e.g., conjunction $\wedge$, disjunction $\lor$). This process generates an explainable causal mediator that approximates an ideal state isolated from confounding effects. Extensive experiments on three large-scale datasets demonstrate CNRE's significant superiority over state-of-the-art baselines, offering multi-level explainability from model design and decision process to recommendation results.
\end{abstract}



\maketitle
\section{Introduction}

In the modern Web ecosystem, understanding personalized user preferences is pivotal. Multi-behavior recommendation (MBR) leverages auxiliary user behaviors to alleviate data sparsity and better capture user preferences \cite{mbht, mpc}, employing architectures ranging from parallel \cite{cvid, MBGMN} to cascading \cite{CRGCN,mbcgcn} methods. Recently, nearly 35\% of MBR models have adopted contrastive learning to boost performance \cite{mbsurvey, HECGCN}. However, this pursuit of accuracy has led to two fundamental issues. First, these methods typically learn a single, uniform representation of user preference, overlooking the distinct preference strengths manifest in a user's behavior chain (e.g., the weak preference of a view-only behavior versus the stronger, escalating intent behind a view→cart progression) \cite{CHMSR}. Second, the reliance on inherently black-box algorithms like contrastive learning sacrifices explainability \cite{CLS}, a cornerstone for building transparent and trustworthy recommender systems.


To enhance the transparency of recommender systems, explainable recommendation has become a crucial research area  \cite{nsexpsurvey}. The dominant paradigm in this field primarily seeks to achieve explainability by incorporating external information. This includes leveraging textual reviews \cite{review1} or attributes \cite{arrtrbute1}, mining explainable reasoning paths on knowledge graphs (KGs) \cite{KG2}.  More recently, the generative capabilities of large language models (LLMs) have also been employed to produce natural language explanations \cite{llm2}, while other studies have introduced causal inference to enhance interpretability \cite{CERS}. A particularly promising technical direction within this paradigm is Neuro-Symbolic (NeSy) integration. By extending symbolic logic to continuous embedding spaces, NeSy synergizes the robust representation learning of neural networks with the explicit reasoning of symbolic systems. This continuous relaxation bridges the gap between perception and reasoning, enabling the derivation of transparent logic rules robust to complex data distributions. This is typically achieved by formulating external information (e.g., attributes or KG triples) as logical predicates composed via logical connectives (e.g., $\wedge$, $\vee$)~\cite{NSS1}. While extensively studied in fields like KG reasoning~\cite{nskg}, NeSy has recently been adapted for recommender systems to provide rule-based explainability~\cite{FENCR}. Table 1 illustrates examples of such logic rules.

\vspace{-1.5mm}
\begin{table}[htb]
 \centering
 \small
 \begin{tabular}{@{}p{2.1cm}|p{6cm}@{}}
  \toprule
  Paradigm & Explainable Logic Rules and Meanings \\
  \midrule
  \multirow{3}{=}{External Information Based} & Rule: $\text{Shop Level}(u) \wedge \text{Item Brand}(i)$ \newline Meaning: Preference attributed to both the user's shop level and the item's brand. \\
  \midrule
  \multirow{3}{=}{Endogenous Logic Based} & Rule: $\text{View}(u, i) \wedge \text{Cart}(u, i) \wedge \text{Buy}(u, i)$ \newline Meaning: Strong preference indicated by the complete user-item behavior chain. \\
  \bottomrule
 \end{tabular}
 \caption{Examples of constructing logic rules and corresponding interpretations.}
 \label{tab:paradigm_comparison}
\end{table}
\vspace{-3.5mm}


However, relying on external information imposes limitations. First, it impedes generalizability and practicality; in multi-behavior scenarios, maintaining extensive attributes across heterogeneous behaviors incurs prohibitive costs, while LLMs introduce hallucination risks. Second, grounding explanations in external information restricts them to being merely declarative. They can clarify what the logic rule is (e.g., attribute A $\wedge$ attribute B), but fail to reveal the model's intrinsic reasoning process—the why and how of its decisions. For instance, KEMB-Rec~\cite{kemb}, the sole existing explainable MBR model, relies on external KGs and black-box contrastive learning~\cite{CLS}, which compromises the faithfulness of its explanations.


To address these challenges, we argue that the user's behavior chain constitutes an endogenous logic: the completeness and type of a preference escalation chain (e.g., view→cart→buy) directly reflect a user's escalating preference strength and correspond to distinct symbolic logic rules (Table 1). However, since interaction data is observational, applying this endogenous logic faces the challenge of spurious correlations caused by confounders \cite{bias}. To this end, Front-door Adjustment is an ideal framework \cite{causalID}, as it effectively isolates the true causal effect, elevating user preference understanding from a correlational to the causal level, while its structured two-step process inherently provides model design explainability.

Building on this, we incorporate causal inference into the Neuro-Symbolic framework to propose a novel \textbf{C}ausal \textbf{N}euro-symbolic \textbf{R}eason\-ing model for \textbf{E}xplainable multi-behavior recommendation (CNRE).  First, the Hierarchical Preference Propagation module mitigates behavior heterogeneity and cross-behavior dependency through behavior-aware parallel encoding and the cascading structure, while its adaptive projection mechanism suppresses confounders from upstream behaviors. Second, the Causal Neuro-Symbolic Reasoning module first models the endogenous logic rule implicit in the user's behavior chain based on its preference strength, and then adaptively dispatches to the corresponding neural-logic reasoning path, such as: (1) Direct processing for strong preferences; (2) Confirmatory conjunction ($\wedge$) inference for medium preferences; and (3) Supplementary disjunction ($\lor$) inference for weak preferences, thereby generating an explainable causal mediator that approximates an ideal state isolated from confounding effects. Crucially, this endogenous approach is highly efficient, avoiding the significant computational overhead of processing external attributes required by models like FENCR \cite{FENCR}. This two-stage process yields robust recommendations grounded in self-contained, multi-level explainability. Our contributions are summarized as follows:

\textbf{Novel paradigm:} We are the first to systematically explore the problem of explainability in multi-behavior recommendation. This paradigm moves beyond the dominant reliance on external information, offering a self-contained and generalizable approach that provides multi-level explainability spanning from model design and decision process to recommendation results.

\textbf{Novel methodology:} CNRE features an adaptive reasoning mechanism that simulates human-like decision-making by translating the strength of a user's behavior chain into distinct neural-logic operations, thereby constructing an explainable causal mediator that approximates an ideal state isolated from confounding effects.

\textbf{Superior performance and explainability:} Extensive experiments on three datasets demonstrate that CNRE outperforms state-of-the-art baselines in recommendation performance, while detailed experimental analyses validate its multi-level explainability.

\section{Related Work}
\subsection{Multi-behavior Recommendation}
Multi-behavior recommendation (MBR) aims to alleviate data sparsity and better profile user preferences by leveraging auxiliary user behaviors \cite{MBCAU,cvid}. Modern MBR methods employ various approaches to encode multi-behavior interactions \cite{mbsurvey}. For instance, parallel modeling approaches  encode different behaviors as independent views to address behavior heterogeneity \cite{mbgcn}. In contrast, cascading modeling approaches  model the user's behavior chain (e.g., view→buy), enhancing the embedding learning for a latter behavior with the embeddings learned from upstream ones to address cross-behavior dependency \cite{CRGCN,cgccn}. To further boost performance, many MBR models have incorporated complex black box techniques like contrastive learning to enhance representation robustness \cite{mbssl}. However, this pursuit of performance has led to a general neglect of model explainability \cite{CLS}. Our work aims to bridge this critical gap by conducting the first systematic exploration of explainability.

\subsection{Explainable Recommendation}
Explainable recommendation aims to enhance system transparency. Methods are broadly categorized as model-intrinsic or post-hoc \cite{nsexpsurvey}. Model-intrinsic approaches often learn preferences from explicit features like reviews \cite{review2} and attributes \cite{fser}, whereas post-hoc methods typically explain black-box models by extracting paths from knowledge graphs \cite{KG1, KG2} or employing LLMs \cite{llm1, llm2}. Recently, causal inference has been introduced to improve robustness \cite{CountER,NCCF}. However, a critical limitation across these paradigms is their heavy reliance on external side information (e.g., KGs, attributes, or text). This dependence acts as a bottleneck, incurring high costs and extraneous noise. In contrast, we achieve multi-level explainability derived solely from endogenous interactions.

\subsection{Neuro-Symbolic Integration}
Neuro-Symbolic (NeSy) integration seeks to unify the robust representation learning capabilities of neural networks with the interpretable logic reasoning of symbolic systems to build trustworthy models \cite{NS2}. 
While heavily explored in fields like knowledge graph reasoning \cite{nskg}, NeSy has recently been adapted for recommender systems to generate explainable recommendations via learned logic rules \cite{nsrs, nsexpsurvey}. 
However, existing paradigms predominantly rely on external knowledge. 
For instance, NS-ICF \cite{NS-ICF} and FENCR \cite{FENCR} derive explainable logic rules from attribute data.  In contrast, we propose a self-contained approach, mining logic rules exclusively from the endogenous logic of multi-behavior interactions.

\section{Preliminaries}
\subsection{Problem Definition}
Let $U$ and $I$ denote the sets of $M$ users and $N$ items, respectively. The user-item interactions are characterized by an ordered chain of behaviors $B = \{b_1, b_2, \dots, b_t\}$, representing increasing preference intensity. Here, the final behavior $b_t$ is designated as the target behavior (e.g., \textit{buy}) that acts as the prediction objective, while the preceding behaviors $\{b_1, \dots, b_{t-1}\}$ serve as auxiliary behaviors. Accordingly, we denote the multi-behavior interaction graphs as $\mathcal{G}=\{\mathcal{G}^{b_1},\mathcal{G}^{b_2},\dots,\mathcal{G}^{b_t}\}$, instantiated as view $\rightarrow$ cart $\rightarrow$ buy for standard scenarios or extended to view $\rightarrow$ collect $\rightarrow$ cart $\rightarrow$ buy specifically for the Tmall dataset. For each graph $\mathcal{G}^b = (\mathcal{V}, \mathcal{E}^b)$, $\mathcal{V}$ denotes the set of nodes (users and items), and $\mathcal{E}^b$ represents the set of edges encoding user interactions under behavior $b$. We construct two separate learnable hypergraphs for users $\mathcal{H}_u^b \in \mathbb{R}^{M \times K}$ and items $\mathcal{H}_i^b \in \mathbb{R}^{N \times K}$ for each behavior $b$, where $K$ is the number of hyperedges. Our task is to compute the probability $\hat{y}_{u,i}^{b_t}$ of the user $u$ interacting with an item $i$ under the target behavior $b_t$.

\subsection{Front-door Adjustment}

In recommender systems, causal inference is essential for mitigating spurious correlations caused by confounders \cite{scm}. Unlike the classic Back-door Adjustment, which requires all confounders to be observable \cite{DCR, HCL}, Front-door Adjustment (FDA) offers a more feasible path. It decomposes the causal effect estimation into two intervenable steps, allowing the model's understanding of user preferences to be deepened from a correlational to a causal level \cite{causalsurvey}. However, FDA requires summing over all possible mediators, which is computationally intractable in the high-dimensional, massive spaces of recommender systems \cite{dccf}. Furthermore, the theoretical do-operator requires an idealized intervention, whereas recommendation models can only perform computation on biased, observational data \cite{causalID}. CNRE treats the Neuro-Symbolic operations on endogenous logic from user behavior chains as a principled simulation of the do-operator, thereby generating an explainable causal mediator $M$ that serves as a tractable, deterministic approximation of an ideal state isolated from confounding effects. We formulate the process of CNRE as the causal graph shown in Figure 1.

\begin{figure}[!ht]
\centering
 \includegraphics[width=\linewidth]{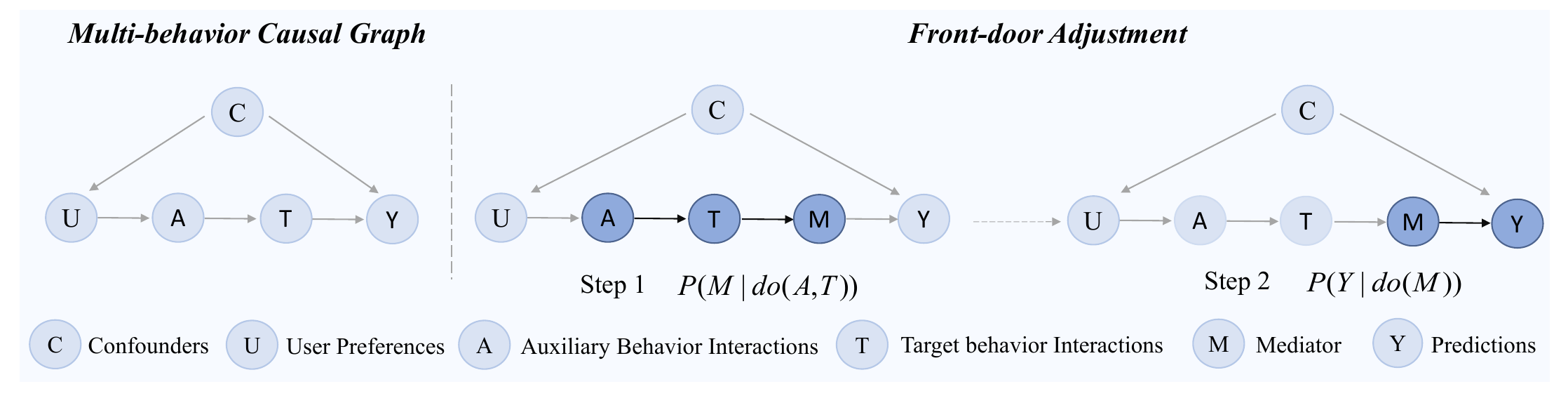}
\caption{Illustration of the causal graph for Front-door Adjustment.}
\label{fig:1}
\end{figure}

\textbf{Edges} $C\rightarrow (U,Y)$ denote that confounders $C$ simultaneously affect both the user's preference $U$ and the prediction $Y$ \cite{bias}; \textbf{Edges}  $U\rightarrow A\rightarrow T$ denote user's preference $U$ first leads to auxiliary behaviors $A$ with weaker preferences, which in turn promotes the generation of the target behavior $T$ with stronger preferences \cite{CRGCN}; \textbf{Edges}  $T\rightarrow Y$ denotes traditional models generating predictions based on the target behavior $T$; \textbf{Edges}  $(A,T)\rightarrow M\rightarrow Y$ denote that model encodes the behavior chain $(A,T)$ into a causal mediator $M$ using causal intervention (do-operation), then generates predictions $Y$, forming the complete front-door path. This decomposition is based on the law of total probability under causal intervention:
\begin{equation}
\begin{split}
    P(Y \mid do(A, T)) = & \sum_{m} P(M = m \mid do(A, T))\cdot \\
    &P(Y \mid do(M = m))
\end{split}
\end{equation}

 \textbf{Step 1:} Effect of the multi-behavior interaction on the mediator $P(M|do(A,T))$. The Hierarchical Preference Propagation module cascading encodes the multi-behavior embeddings for both users and items ($A\rightarrow T$). During this process, the adaptive projection mechanism proactively suppresses confounding signals from upstream behaviors. Subsequently, neural-logic operations encode the mediator $(A,T)\rightarrow M$ by performing a computational causal intervention ($do(A,T)$). This process satisfies the first FDA condition as $M$ is computed solely from the embeddings of $A$ and $T$, structurally blocking any back-door path from confounders $C$ to $M$. Thus, the intervention effect can be unbiasedly identified $P(M|A,T)$ \cite{scm}.

\textbf{Step 2:} Effect of the mediator on the recommendation result. The second step of FDA is to estimate the causal effect of the mediator $M$ on the final recommendation by computing $P(Y|do(M))$. To achieve this, our model relies solely on the mediator $M$ for the final prediction $(M \rightarrow Y)$. This architectural choice ensures that $M$ intercepts all directed paths from $(A,T)$ to $Y$, thereby satisfying a key condition of FDA. By learning this direct mapping from the disentangled mediator $M$ to the outcome $Y$, the model is trained to approximate the true causal effect and is structurally isolated from the confounding path $M$ ← $(A,T)$ ← $U$ → $Y$. Our designed two-stage reasoning process thus structurally fulfills the requirements of FDA, providing model design explainability.
\begin{figure*}[!ht]
\centering
 \includegraphics[width=\linewidth]{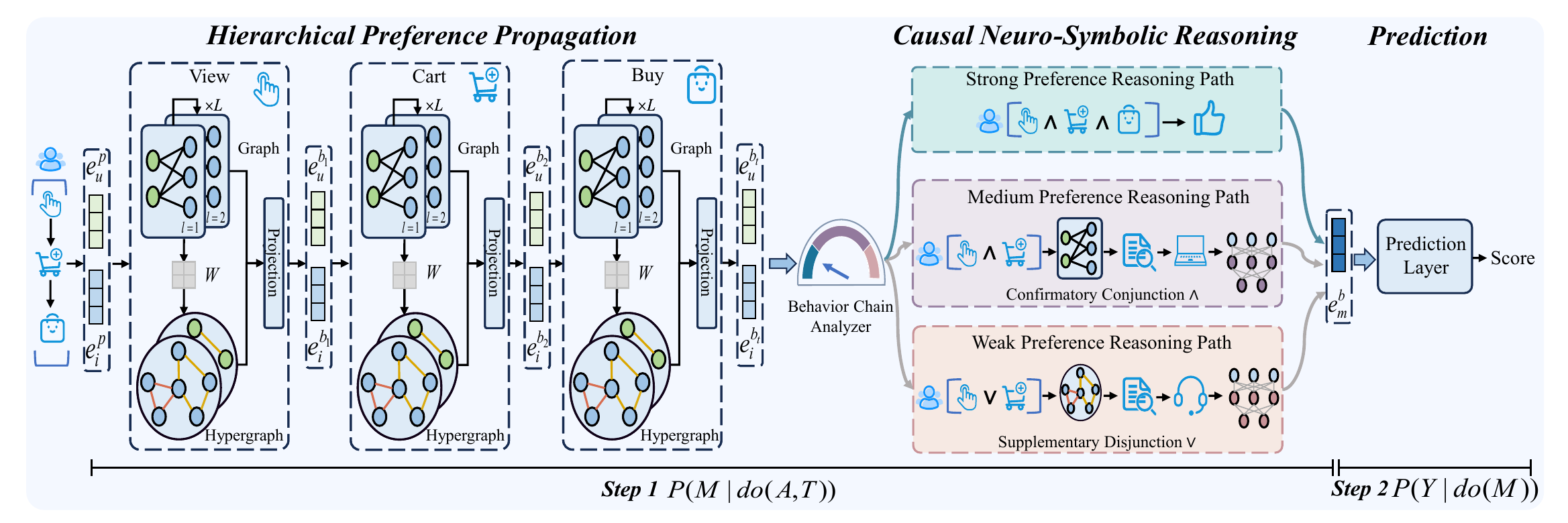}
\caption{Illustration of our proposed CNRE model.}
\label{fig:1}
\end{figure*}

\section{Proposed Method}
Figure 2 illustrates the overall framework, which comprises three core modules: (1) Hierarchical Preference Propagation; (2) Causal Neuro-Symbolic Reasoning; (3) Prediction.

\subsection{Hierarchical Preference Propagation}
This module is designed to address three challenges: the cross-behavior dependency arising from preference escalation, the inherent heterogeneity within each behavior, and the confounding effects from massive upstream behaviors. By systematically modeling this progression, the module captures personalized user preferences, thereby providing high-quality embedding representations for the subsequent Causal Neuro-Symbolic Reasoning.
\subsubsection{Intrinsic Preference Learning} To capture a user's intrinsic preferences across multiple behaviors and to provide a robust initial embedding for the cascading encoding. We first construct a unified graph $\mathcal{G}$ that contains all behavior interactions and apply LightGCN \cite{lightgcn} for L-layer iterative aggregation of embeddings: 
\begin{equation}
e_u^{p,(l)} = \sum_{i \in \mathcal{N}_u} \frac{1}{\sqrt{|\mathcal{N}_u||\mathcal{N}_i|}} e_i^{p,(l-1)}
\label{eq:layer_aggregation}
\end{equation}
where $\mathcal{N}_u$ denotes the set of items that user $u$ has interacted with, $l$ denotes the layer. Final intrinsic preference embeddings for users $e_u^{p}$ and items $e_i^{p}$ are generated by summing the embeddings from each layer: $e_{(u,i)}^{p}=\sum_{l=0}^{L}e_{(u,i)}^{p,(l)}$, which will be used to enhance the representation learning for the first behavior $e_{(u,i)}^{p} \rightarrow e_{(u,i)}^{b_1}$ in the behavior chain.

\subsubsection{Behavior-aware Parallel Encoding} We first determine the optimal cascading order of behaviors for different datasets by analyzing the behavior conversion rates. To address behavior heterogeneity, we then perform two parallel graph encoding operations within each behavior $b$ to capture collaborative and semantic signals.

Collaborative signal encoding: On the behavior-specific user-item interaction graph $\mathcal{G}^{b}$, we apply LightGCN by Eq.\eqref{eq:layer_aggregation} for L-layer embedding aggregation to capture the collaborative signal. By summing the embeddings from each layer's output $e_{col}^b=\sum_{l=0}^{L}e_{col}^{b,(l)} $, we obtain the behavior-specific user and item collaborative embeddings $e_{u,col}^{b}, e_{i,col}^{b}$.

High-order semantic encoding: This component aims to address the limitations of collaborative signals by capturing associations beyond pairwise relationships, a crucial step for modeling behavior heterogeneity. To this end, we introduce the learnable hypergraph structure to uncover implicit semantic structures, where the hyperedge incidence matrices are generated by projecting the collaborative embeddings through learnable matrices $\mathbf{W}_{(u,i),hyp}^b$:

\begin{equation}
\mathcal{H}_u^b = e_{u,col}^{b} \mathbf{W}_{u,hyp}^b, \mathcal{H}_i^b = e_{i,col}^{b} \mathbf{W}_{i,hyp}^b
\end{equation}

Then capture high-order correlations via hypergraph convolution \cite{hgcn}. This is implemented by constructing affinity matrices from the incidence matrices, which are then used to propagate information and yield the user and item semantic embeddings:

\begin{equation}
e_{u,sem}^{b} = (\mathcal{H}_u^b (\mathcal{H}_u^b)^\top) e_{u,col}^{b} ,e_{i,sem}^{b} = (\mathcal{H}_i^b (\mathcal{H}_i^b)^\top) e_{i,col}^{b}
\end{equation}

\subsubsection{Cascading Adaptive Projection} To mitigate the propagation of confounders from upstream auxiliary behaviors, we employ an adaptive projection within each behavior layer. This projection calibrates the high-order semantic embeddings against the stable collaborative signals captured by the interaction graph. By filtering out inconsistent components, this operation suppresses confounding signals to ensure cross-view embedding consistency while preserving core user preferences:
\begin{equation}
\hat{e}_{sem}^{b} = \frac{e_{col}^{b} \cdot e_{sem}^{b}}{\|e_{col}^{b}\|^2_2 + \epsilon} e_{col}^{b}
\end{equation}

The upstream behavior embeddings $e^{b-1}$, collaborative embeddings $e_{col}^b$, and the calibrated semantic embeddings $\hat{e}_{sem}^{b}$ from the current behavior $b$ are then aggregated:

\begin{equation}
e^b = e^{b-1} + e_{col}^b + \hat{e}_{sem}^{b}
\end{equation}

The aggregated embeddings (for both users $u$ and items $i$) from the current behavior $b$ serve as the initialization for the collaborative signal embeddings of the subsequent behavior $b+1$:
\begin{equation}
e_{col}^{b+1,(l=0)}=e^b
\end{equation}

This process enhances representation learning on the user-item interaction graph, facilitating a cascading encoding scheme that effectively resolves cross-behavior dependencies.

\subsection{Causal Neuro-Symbolic Reasoning}
CNRE treats each behavior type as a binary predicate. By analyzing the user's behavior chain for the target item, it adaptively constructs logic rules for varying preference strengths and dispatches to one of three distinct reasoning paths for a Neuro-Symbolic operation.

\subsubsection{Strong Preference Reasoning Path} When the user's behavior chain is complete, it instantiates the strongest logic rule (e.g., view $\wedge$ cart $\wedge$ buy), signifying a clear progressive preference despite the presence of confounders in auxiliary behaviors. The causal mediator $e_m^b$ is generated by concatenating the final behavior-specific embeddings of the user and item from the Hierarchical Preference Propagation. Notably, these embeddings have already been calibrated by the adaptive projection, which effectively functions as a computational causal intervention to block upstream confounders. This ensures that even this reasoning path adheres to the independence assumptions of the Front-door Adjustment.

\subsubsection{Medium Preference Reasoning Path} When the user's behavior chain is incomplete but contains multiple behaviors, it forms the ambiguous logic rule (e.g., view $\wedge$ cart), and their decision intent is uncertain, which  triggers the confirmatory conjunction ($\wedge$) inference. To account for behavior heterogeneity and cross-behavior dependency within the behavior chain, this process is initiated by calculating purchase confidence score using the embeddings of the final behavior $b$, in behavior chain (e.g., cart). If this score falls below the threshold $\tau$, it indicates the preference signal is insufficient for a final decision. To enhance the user's preference, we retrieve the most relevant collaborative embedding for the target item $i$ from the final behavior's $b$ item embedding space:

\begin{equation}
S_{col}^b = \text{ANNS}(e_{i,col}^b, \mathcal{V}_b, N_c)
\label{eq:anns}
\end{equation}
where $\text{ANNS}$ is the approximate nearest neighbor search, $\mathcal{V}_b$ is the search space for behavior $b$, $N_c$ is hyperparameter. Following LINN \cite{nlr}, we use an MLP as a logical operator to perform the neural conjunction operation, generating the causal mediator $e_m^b$:

\begin{equation}
e_m^b = \mathbf{W}_2 \sigma(\mathbf{W}_1 (e_u^b \oplus e_{i,col}^b \oplus S_{col}^b) + \mathbf{b}_1) +  \mathbf{b}_2
\label{eq:mlp}
\end{equation}
where $\oplus$ denotes the concat, $e_u^b$ denotes the behavior-specific user embedding, $\mathbf{W},\mathbf{b}$ denotes weight matrices and bias.

\subsubsection{Weak Preference Reasoning Path} When user exhibits only a single auxiliary behavior (e.g., view), the preference signal is weak and collaborative information is unreliable due to data sparsity. Therefore, we perform the supplementary disjunction ($\lor$) inference. Without calculating the confidence score, we leverage the semantic embeddings $e_{i,sem}^{b}$, which contain high-order semantic correlations mined from the learnable hypergraph, to enhance the user's preference. We retrieve the most relevant semantic embedding $S_{sem}^b$ for the target item i from the hypergraph's semantic space for the final behavior $b$, in behavior chain (e.g., view):
\begin{equation}
S_{sem}^b = \text{ANNS}(e_{i,sem}^b, \mathcal{V}_b, N_c)
\label{eq:anns}
\end{equation}

A structurally identical but parametrically independent logical operator then performs a neural disjunction operation to generate the causal mediator $e_m^b$:
\begin{equation}
e_m^b = \tilde{\mathbf{W}_2} \sigma(\tilde{\mathbf{W}_1} (e_u^b \oplus e_{i,sem}^b \oplus S_{sem}^b) + \tilde{\mathbf{b}}_1) + \tilde{\mathbf{b}}_2
\end{equation}


\subsection{Prediction}
The Prediction Module fulfills the second step of FDA, aiming to estimate the causal effect $P(Y|do(M))$. Unlike traditional dot-product methods that require the interaction of separate user and item embeddings, this step necessitates predicting the outcome based solely on the unified mediator representation. Accordingly, we utilize an MLP \cite{neumf} to derive predictions directly from the target-specific $e_m^{b_t}$:

\begin{equation}
\hat{y}_{u,i}^{b_t} = \sigma ( \mathbf{W}^\top ( \text{MLP} (e_m^{b_t} ) )+ \mathbf{b} )
\end{equation}
where $\sigma$ is the activation function. We use the BPR \cite{bpr} to maximize the predicted score margin between observed positive pairs $(u,i)$ and negative pairs $(u,j)$:

\begin{equation}
\mathcal{L}_{BPR}^b = \sum_{(u,i,j) \in \mathcal{D}} -\log\sigma(\hat{y}_{u,i}^b - \hat{y}_{u,j}^b)
\end{equation}

We adopt a multi-task learning strategy to better learn the parameters from multiple behaviors and use L2 regularization to enhance the model's generalization ability:
\begin{equation}
\mathcal{L} = \sum_{b=1}^{B} \mathcal{L}_{BPR}^b + \lambda \|\Theta\|_2^2
\end{equation}

\section{Experiments}

In this section, we conduct comprehensive experiments to evaluate CNRE, including: overall performance comparison against different baselines (RQ1), ablation study on key components (RQ2), multi-level explainability analysis (RQ3), analysis of the model's robustness and hyperparameter sensitivity (RQ4).

\begin{table}[h]
\begin{center}
\small
\caption{Statistics of the experimental datasets.}
\label{tab:dataset_statistics_final}
\setlength{\tabcolsep}{6pt}
\begin{tabular}{@{}ccccc@{}}
\toprule
\textbf{Dataset} & \textbf{Users} & \textbf{Items} & \textbf{Interactions} & \textbf{Behavior Type} \\
\midrule
Beibei & 21,716 & 7,977 & 3,338,068 & View, Cart, Buy \\
Taobao & 48,749 & 39,493 & 1,952,931 & View, Cart, Buy \\
Tmall & 15,449 & 11,953 & 1,395,273 & View, Collect, Cart, Buy \\
\bottomrule
\end{tabular}
\end{center}
\end{table}

\begin{table*}[t]
\centering
\small
\caption{Overall performance in terms of HR@K and NDCG@K on the three datasets (abbreviated as H@K and N@K). Bold numbers indicate the best performance, "*" indicates the statistical significance over the best baseline using a t-test with $p$ < 0.05.}
\setlength{\tabcolsep}{4pt} 
\begin{tabular}{@{}c|c|cccc|cccc|cccc@{}}
\toprule
\multirow{2}{*}{\textbf{Type}} & \multirow{2}{*}{\textbf{Method}} & \multicolumn{4}{c|}{\textbf{Beibei}} & \multicolumn{4}{c|}{\textbf{Taobao}} & \multicolumn{4}{c}{\textbf{Tmall }} \\
\cmidrule(lr){3-6} \cmidrule(lr){7-10} \cmidrule(lr){11-14}
& & H@10 & N@10 & H@50 & N@50 & H@10 & N@10 & H@50 & N@50 & H@10 & N@10 & H@50 & N@50 \\
\midrule
\multirow{1}{*}{Single-behavior} & NeuMF      & 0.0161 & 0.0088 & 0.0348 & 0.0112 & 0.0085 & 0.0037 & 0.0385 & 0.0154 & 0.0129 & 0.0062 & 0.0297 & 0.0109 \\
\midrule
\multirow{6}{*}{Multi-behavior} & MBGCN      & 0.0393 & 0.0219 & 0.0847 & 0.0315 & 0.0433 & 0.0244 & 0.0938 & 0.0351 & 0.0487 & 0.0232 & 0.1062 & 0.0353 \\
& MBSSL      & 0.0610 & 0.0344 & 0.1136 & 0.0595 & 0.0847 & 0.0420 & 0.1712 & 0.0639 & 0.0734 & 0.0416 & 0.1541 & 0.0645 \\
& CRGCN      & 0.0389 & 0.0271 & 0.0841 & 0.0388 & 0.0796 & 0.0384 & 0.1721 & 0.0554 & 0.0715 & 0.0427 & 0.1525 & 0.0669 \\
& MB-CGCN    & 0.0507 & 0.0325 & 0.1096 & 0.0469 & 0.1165 & 0.0626 & 0.2517 & 0.0898 & 0.1044 & 0.0592 & 0.2088 & 0.0858 \\
& BCIPM      & 0.0743 & 0.0353 & 0.1608 & 0.0573 & 0.1361 & 0.0768 & 0.2942 & 0.1193 & 0.1252 & 0.0738 & 0.2379 & 0.0994 \\
\midrule
\multirow{2}{*}{Causal} & CausalD   & 0.0650 & 0.0312 & 0.1365 & 0.0465 & 0.0924 & 0.0471 & 0.1932 & 0.0705 & 0.0783 & 0.0461 & 0.1638 & 0.0690 \\
& DCCF      & -  & -  & -  & -  & 0.0952 & 0.0486 & 0.1995 & 0.0723 & 0.0807 & 0.0476 & 0.1682 & 0.0705 \\

\midrule
\multirow{2}{*}{Explainable} & NS-ICF     & -      & -      & -      & -      & 0.0784 & 0.0339 & 0.1686 & 0.0486 & 0.0682 & 0.0341 & 0.1488 & 0.0563 \\
& FENCR      & -      & -      & -      & -      & 0.0893 & 0.0445 & 0.1928 & 0.0638 & 0.0760 & 0.0452 & 0.1614 & 0.0697 \\
\midrule
\multirow{1}{*}{\textbf{Ours}} & \textbf{CNRE} & \textbf{0.1074$^*$} & \textbf{0.0523$^*$} & \textbf{0.2310$^*$} & \textbf{0.0758$^*$} & \textbf{0.1592$^*$} & \textbf{0.0976$^*$} & \textbf{0.3435$^*$} & \textbf{0.1398$^*$} & \textbf{0.1385$^*$} & \textbf{0.0827$^*$} & \textbf{0.2936$^*$} & \textbf{0.1158$^*$} \\
\bottomrule
\end{tabular}
\label{tab:overall_performance_full}
\end{table*}

\subsection{Datasets}
We validate the effectiveness of our proposed model on three widely-used, public multi-behavior datasets. The Beibei dataset \cite{mbcgcn}, sourced from a leading e-commerce platform for maternal and infant products, includes three interaction types: page view (view), add-to-cart (cart), and buy. The Taobao dataset \cite{taobao}, collected from one of China's largest e-commerce platforms, contains the same set of behaviors. Similarly, the Tmall dataset \cite{bcipm} encompasses four behavior types: view, collect, cart, and buy. Detailed statistics for all datasets are summarized in Table 2.

\subsection{Baselines}
We compare our method with the following three types of recommendation models: \textbf{Single-behavior Model} like NeuMF \cite{neumf} a neural collaborative filtering approach. \textbf{Multi-behavior Models} including parallel MBGCN \cite{mbgcn} , cascading (e.g., CRGCN \cite{CRGCN}, MB-CGCN \cite{mbcgcn}), and contrastive learning-based (e.g., MBSSL \cite{mbssl}) architectures, as well as models that learn behavior contextualized item preferences (e.g., BCIPM \cite{bcipm}). \textbf{Causal Inference Models} including DCCF \cite{dccf}, which employs FDA on item features to mitigate confounders, and  CausalD \cite{causalID}, which utilizes causal distillation to alleviate performance heterogeneity. \textbf{Explainable Neuro-Symbolic Models} include NS-ICF\cite{NS-ICF}, which learns rules from attributes using a three-tower architecture, and FENCR \cite{FENCR}, which enhances attribute-based rules with feature embeddings.


\subsection{Overall Performance (RQ1)}
We performed extensive experiments to make recommendations on three public datasets. From the results in Table 3, we summarize the following key observations: Consistent with prior work, all multi-behavior models significantly outperform the single-behavior baseline, confirming that leveraging multiple types of behavior provides a more comprehensive understanding of user preferences.

Compared to other multi-behavior models, CNRE's hybrid architecture of cascading (CRGCN and MB-CGCN) and parallel (MBGCN) encoding systematically mitigates behavior heterogeneity and dependency. Unlike black-box contrastive learning models such as MBSSL and models focusing on specific behaviors like BCIPM, CNRE models the entire preference escalation chain through a two-stage Front-door Adjustment, thereby achieving superior performance while providing multi-level explainability. This also represents a paradigm shift compared to Neuro-Symbolic models like NS-ICF and FENCR. Instead of relying on external attribute information to construct declarative rules, CNRE mines endogenous logic from the behaviors themselves to provide a deeper, procedural explanation for its reasoning process, while simultaneously improving performance. Furthermore, while advanced causal models like DCCF and CausalD also aim to learn purer user preferences, their focus is on debiasing a single behavior signal to obtain performance gains, lacking both the rich information from auxiliary behaviors and the intrinsic explainability of CNRE's reasoning process. This independence from external information grants CNRE greater generality, allowing it to operate on datasets like Beibei, on which DCCF, NS-ICF and FENCR cannot run due to missing required attributes.



\begin{table}[t]
\centering
\small
\caption{Ablation study of key components on CNRE.}
\setlength{\tabcolsep}{5.5pt} 
\begin{tabular}{@{}c|cc|cc|cc@{}}
\toprule
\multirow{2}{*}{\textbf{Methods}} & \multicolumn{2}{c|}{\textbf{Beibei}} & \multicolumn{2}{c|}{\textbf{Taobao}} & \multicolumn{2}{c}{\textbf{Tmall } } \\
\cmidrule(lr){2-3} \cmidrule(lr){4-5} \cmidrule(lr){6-7}
& H@10 & N@10 & H@10 & N@10 & H@10 & N@10 \\
\midrule
\multicolumn{7}{c}{\textbf{Hierarchical Preference Propagation Module}} \\
\midrule
\quad w/o HPP & 0.0846 & 0.0442 & 0.1237 & 0.0898 & 0.1080 & 0.0728 \\
\quad w/o PAR & 0.0954 & 0.0485 & 0.1392 & 0.0936 & 0.1191 & 0.0769 \\
\quad w/o PRJ & 0.1025 & 0.0504 & 0.1454 & 0.0955 & 0.1260 & 0.0802 \\
\midrule
\multicolumn{7}{c}{\textbf{Causal Neuro-Symbolic Reasoning Module}} \\
\midrule
\quad w/o REA & 0.0930 & 0.0469 & 0.1355 & 0.0914 & 0.1177 & 0.0752 \\
\quad w/o CNJ & 0.0985 & 0.0488 & 0.1451 & 0.0935 & 0.1261 & 0.0786 \\
\quad w/o DSJ & 0.1012 & 0.0502 & 0.1423 & 0.0928 & 0.1220 & 0.0782 \\
\midrule
\textbf{CNRE} & \textbf{0.1074} & \textbf{0.0523} & \textbf{0.1592} & \textbf{0.0976} & \textbf{0.1385} & \textbf{0.0827} \\
\bottomrule
\end{tabular}
\label{tab:ablation_study}
\end{table}

\subsection{Ablation Study (RQ2)}
To validate the contribution of each key component in the CNRE framework, we conducted an ablation study with the following variants: (1) \textbf{w/o HPP:} removes the Hierarchical Preference Propagation module and replaces it with a simple parallel GCN encoder; (2) \textbf{w/o PAR:} removes the parallel hypergraph enhancement; (3) \textbf{w/o PRJ:} removes the adaptive projection mechanism;  (4) \textbf{w/o REA:} removes the core Causal Neuro-Symbolic Reasoning module; (5) \textbf{w/o CNJ:} removes the confirmatory conjunction inference path; (6) \textbf{w/o DSJ:} removes the supplementary disjunction inference path. 

Table 4 validates the effectiveness of our Hierarchical Preference Propagation module. Specifically, the decline in w/o HPP proves that simple parallel encoding fails to capture the progressive escalation of user intent across the behavior chain. Similarly, the results for w/o PAR and w/o PRJ highlight the critical role of parallel hypergraph encoding in capturing high-order correlations amidst behavior heterogeneity, and the adaptive projection in effectively disentangling true preference signals from upstream confounders.

Removing the entire reasoning module (w/o REA) precipitates a significant performance decline and, crucially, results in a complete loss of CNRE's multi-level explainability.  The degradation of w/o CNJ is particularly pronounced on the dense Beibei dataset, highlighting the confirmatory conjunction's effectiveness in filtering noise and solidifying ambiguous, medium-preference signals (e.g., $View \wedge Cart$). Conversely, the performance drop for w/o DSJ is more significant on the sparser Taobao dataset. This demonstrates that the supplementary disjunction inference successfully alleviates data sparsity by retrieving high-order semantic embedding to infer potential interests when explicit signals are weak.


\subsection{Multi-level Explainability Analysis (RQ3)}
By modeling the endogenous logic of multi-behaviors, CNRE provides a multi-level explainability framework. Consistent with the standard practice in multi-behavior recommendation research that utilizes public, offline datasets, conducting a large-scale user study was infeasible for this work. Therefore, drawing upon the analytical paradigms of existing explainability research, this section presents a comprehensive and in-depth analysis of CNRE's explainability.

\subsubsection{Model Design Explainability}

The design-level explainability of CNRE is detailed in Section 3.2 and depicted in the Figure 1. The model's architecture decomposes the reasoning process into the two requisite stages of FDA: first, estimating the effect of the behavior chain on the causal mediator $P(M|do(A,T))$, and second, estimating the effect of the mediator on the final prediction $P(Y|do(M))$. This two-stage process ensures that CNRE's architecture is not an opaque "black box," but is instead inherently transparent and possesses the theoretical explainability provided by FDA.

\subsubsection{Decision Process Explainability}
The explainability of our decision process stems from a fundamental shift from the implicit representation learning of cascading models (e.g., CRGCN, MB-CGCN) to an explicit, adaptive reasoning mechanism. Specifically, CNRE employs three distinct reasoning paths based on the preference strength manifested in the user's behavior chain: supplementary disjunction inference to alleviate weak preferences, confirmatory conjunction inference to validate uncertain preferences, and direct recommendation for strong buy preferences. To quantitatively validate this process, we analyze the activation distribution of these three paths on the Taobao dataset, as shown in Figure 3(a). This figure corrects a visualization rounding artifact present in the conference proceedings version.

This three-path division is a principled design choice that mirrors the e-commerce conversion funnel. For instance, the weak preference path addresses scenarios where the user's behavior signal is singular and highly uncertain, while the medium preference path handles cases of composite behaviors where intent is enhanced but the decision remains incomplete. This logic rule-based design is crucial for procedural explainability, which would be lost if these paths were learned adaptively by another black-box model.

\begin{figure}[htb]
    \centering 

    \begin{subfigure}[b]{0.48\linewidth}
        \centering
        \includegraphics[width=\linewidth]{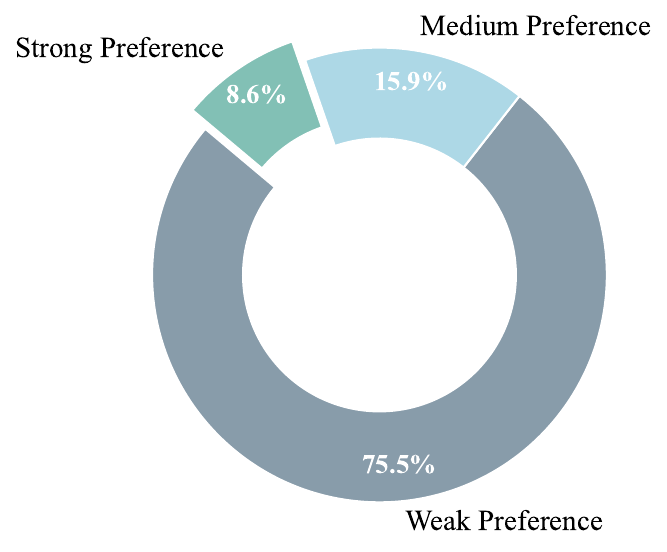}
        \subcaption{Reasoning path distribution}
        \label{fig:sub1}
    \end{subfigure}
    \hfill 
    \begin{subfigure}[b]{0.47\linewidth}
        \centering
        \includegraphics[width=\linewidth]{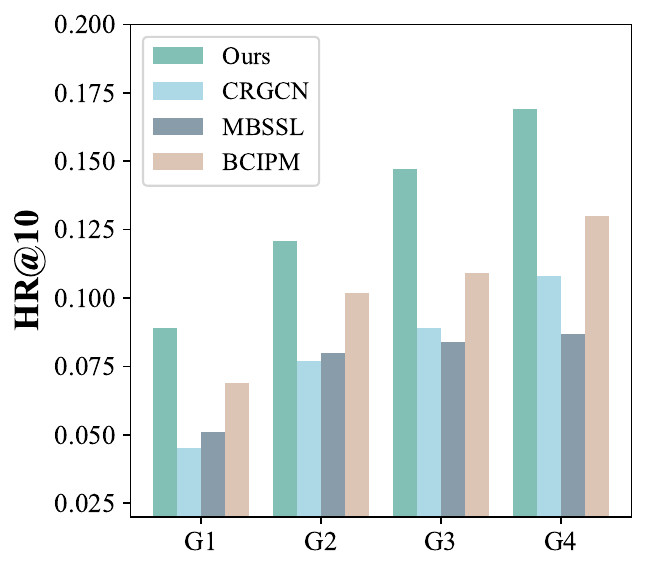}
        \subcaption{Different interaction sparsity}
        \label{fig:sub2}
    \end{subfigure}

    \caption{Analysis of decision process explainability.}
    \label{fig:main_figure}
\end{figure}

To validate the adaptiveness of our reasoning process, we analyzed user groups with varying sparsity (G1-G4). CNRE consistently outperforms baselines across all densities (Figure 3(b)). The model adaptively employs supplementary disjunction in low-density scenarios (G1, G2) to enhance weak signals, while shifting to confirmatory conjunction in high-density scenarios (G3, G4) to validate uncertain preferences. This dynamic path selection makes the model's decision process transparent and explainable.

\subsubsection{Recommendation Result Explainability}
Aligning with Neuro-Symbolic surveys~\cite{NSS1} that identify the explicit tracking of inference steps as a core source of explainability, we present case studies from the Taobao dataset in Table 5. We validate CNRE's transparency via counterfactual analysis, examining how perturbing the behavior chain impacts the model's inference path and results.

\begin{table*}[t]
\centering
\label{tab:case_study_final}
\small
\setlength{\tabcolsep}{7pt}
\caption{Case study and counterfactual analysis of CNRE's recommendation result explainability. * denotes a counterfactual analysis of the corresponding base case. Bolded values in the counterfactual rows highlight deviations from the original case.}
\begin{tabular}{@{} p{1.0cm}  p{1.1cm}  p{3.2cm} p{3.2cm}  p{7.3cm} @{}}
\toprule
\textbf{Case} & \textbf{Item ID} & \textbf{Behavior Chain} & \textbf{Inference Path} & \textbf{Inferred Rationale \& Recommendation} \\
\midrule
\multirow{3}{=}{\#1} & \multirow{3}{=}{4350 } & \multirow{3}{=}{view=1, cart=1, buy=0} & \multirow{3}{=}{Medium Preference; Conjunction Inference $\wedge$} & 
Initial Confidence $<$ $\tau$ →  Retrieve Collaborative Embedding (item 32039) → Neural Conjunction Operation $\wedge$ → Causal Mediator → Final Recommendation (item 12226, score 0.713).
 \\

\midrule 

\multirow{3}{=}{\#2} & \multirow{3}{=}{22470} & \multirow{3}{=}{view=0, cart=1, buy=0} & \multirow{3}{=}{Weak Preference; Disjunction Inference $\lor$} & 
Without Confidence Calc → Retrieve Semantic Embedding (item 32755) → Neural Disjunction Operation $\lor$ → Causal Mediator → Final Recommendation (item 21517, score 0.662). \\
\midrule
\multirow{3}{=}{\#1*} & \multirow{3}{=}{4350} & \multirow{3}{=}{view=1,  \textbf{cart=0}, buy=0} & \multirow{3}{=}{\textbf{Weak Preference; Disjunction Inference $\lor$}} & 
When 'cart' is removed, the path switches to \textbf{weak inference}. Skip Confidence Calc → Retrieve Semantic Embedding (item \textbf{34516}) → Final Recommendation (item \textbf{956}, score \textbf{0.652}). \\
\midrule
\multirow{3}{=}{\#2*} & \multirow{3}{=}{22470} & \multirow{3}{=}{ \textbf{view=1}, cart=1, buy=0} & \multirow{3}{=}{\textbf{Medium Preference; Conjunction Inference $\wedge$}} & 
When 'view' is added, the path upgrades to \textbf{medium inference}. Initial Confidence $<$ $\tau$ → Retrieve Collaborative Embedding (item \textbf{19515}) → Final Recommendation (item \textbf{7180}, score \textbf{0.595}). \\

\bottomrule
\end{tabular}
\end{table*}

In Case \#1, for a medium-strength chain (view=1, cart=1), the model adaptively generates the logic rule view $\wedge$ cart and activates the confirmatory conjunction inference ($\wedge$) path. The corresponding counterfactual analysis (\#1*) shows that removing the 'cart' behavior prompts a logical downgrade to the weak preference path (disjunction $\lor$), strongly demonstrating the explanation's causal consistency. In Case \#2, for a weak signal (view=1, cart=0) where collaborative information can be unreliable, the model employs the supplementary disjunction inference ($\lor$). This path retrieves semantic embeddings from the hypergraph to enhance the weak preference, a crucial function in data-sparse scenarios. Its counterfactual analysis (\#2*) reveals a profound, non-monotonic reasoning: while adding the view behavior upgrades the path as expected, the recommendation score counter-intuitively decreases. This phenomenon arises because the reasoning process is inherently context-sensitive: the path transition reconfigures the logical expression, and the newly incorporated collaborative embedding may encode negative latent patterns, thereby justifying a more conservative and calibrated prediction.

Notably, CNRE does not utilize temporal information; thus, the cascading encoding process inherently precludes interactions that violate the behavior chain (e.g., cart→view). We directly activate the strong preference path for skip-behavior chains (e.g., view→buy). Even though upstream auxiliary behaviors are affected by confounders, the buy behavior, as the ground truth, embodies the user's strongest preference, and prioritizing it is also a widely-recognized strategy in the recommender system domain.

\subsubsection{Efficiency Evaluation of Explainability}
Unlike post-hoc models such as CausalX \cite{llm1} and LLMXRec \cite{llm2} that incur additional overhead, CNRE's explainability is intrinsic. Its reasoning path is embedded in the forward pass, yielding explanations that are structurally faithful and computationally efficient.

To theoretically quantify this efficiency, we further analyze CNRE's time complexity. Let $L$ be the number of GCN layers, $d$ is the embedding dimension, $E_{total}$ is the total number of interactions, and $b_s$ the batch size. CNRE's total complexity is $O(L \cdot E_{total} \cdot d + b_s \cdot d^2)$. This is comparable to BCIPM $O(L \cdot E_{total} \cdot d + b_s \cdot N_c \cdot d^2)$, but our reasoning cost is superior as it does not scale with the neighborhood size $N_c$. The advantage is most pronounced when compared to the Neuro-Symbolic model FENCR, whose complexity is $O(b_s \cdot N_{rules} \cdot A \cdot d^2)$. FENCR's reasoning cost scales linearly with its number of predefined rules ($N_{rules}$) and external attributes ($A$), whereas CNRE's core reasoning module $O(b_s \cdot d^2)$ is independent of these factors. This reliance on external information makes FENCR computationally expensive and difficult to scale, especially in multi-behavior contexts where managing distinct attribute sets for each behavior could lead to a combinatorial explosion in complexity.

\subsection{Robustness and Hyperparameter Sensitivity Analysis (RQ4)}
We simulated cold-start scenarios on the Taobao dataset by progressively removing 10\% to 50\% of historical interactions for 50\% of the users. As shown in Figure 4a, while all models' performance declined with increasing data sparsity, CNRE consistently exhibited the smallest degradation and significantly outperformed all baselines. This is attributed to our hybrid architecture, which addresses behavior heterogeneity and cross-behavior dependency, and our causal reasoning module, which enhances preferences at varying strengths by constructing logic rules and reasoning.

\begin{figure}[htb]
    \centering 

    \begin{subfigure}[b]{0.48\linewidth}
        \centering
        \includegraphics[width=\linewidth]{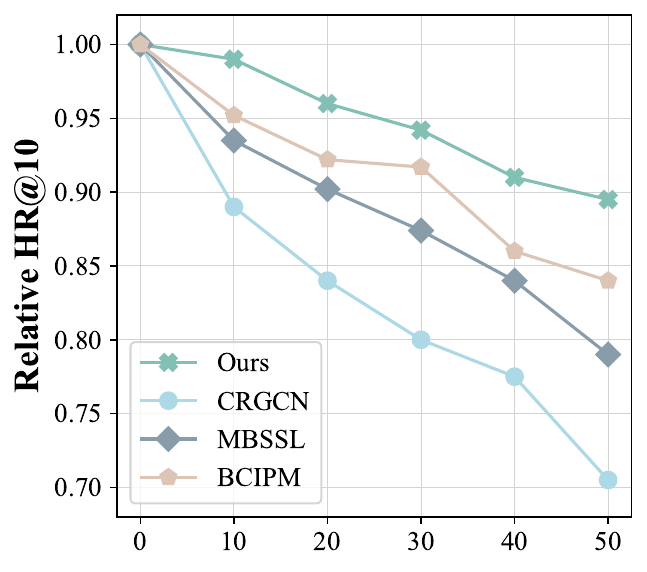}
        \subcaption{Cold-start scenarios}
        \label{fig:sub1}
    \end{subfigure}
    \hfill 
    \begin{subfigure}[b]{0.49\linewidth}
        \centering
        \includegraphics[width=\linewidth]{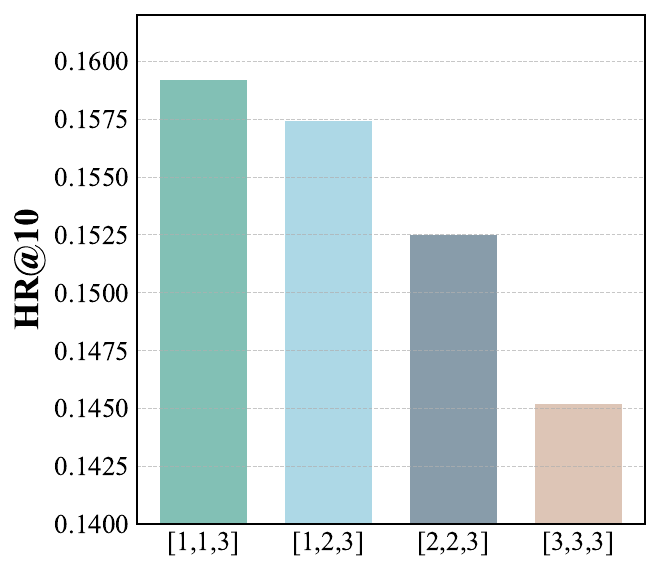}
        \subcaption{Cascading GCN Layers}
        \label{fig:sub2}
    \end{subfigure}

    \caption{Robustness and cascading effectiveness analysis.}
    \label{fig:main_figure}
\end{figure}

As the data distribution and weight of different behaviors vary,  setting the optimal number of graph convolution layers for each behavior in the chain is crucial for the cascading method to encode a user's progressive preferences. Our parameter analysis (Figure 4b) shows that the combination [1,1,3] achieves the best performance. This is because upstream behaviors (view, cart) are dense and more susceptible to confounders. For the downstream target behavior, which is sparse but carries the strongest weight; therefore, using the 3-layer GCN leads to performance improvements.

\section{Conclusions and Future Work}


In this paper, we propose a novel Causal Neuro-Symbolic reasoning model CNRE. CNRE simulates a human-like reasoning process by adaptively dispatching to distinct neural-logic paths based on the preference strength inferred from a user's behavior chain. By modeling and reasoning on these endogenous logic rules, our model achieves significant performance gains in recommendation accuracy and provides multi-level explainability, as demonstrated by extensive experiments on three real-world datasets. Acknowledging the lack of recognized quantitative metrics for explainable recommendation and that robust user studies require extensive multidisciplinary knowledge, our future work will focus on designing a human-centric, quantifiable evaluation framework and conducting large-scale user studies to validate our model's real-world utility and alignment with human cognitive reasoning.


\bibliographystyle{ACM-Reference-Format}
\balance
\bibliography{sample-base}

\appendix
\section{Implementation Details}
For all baselines, we follow the hyperparameter settings recommended in their original papers. Our CNRE model is implemented in PyTorch and trained end-to-end using the Adam optimizer. We set the batch size to 1024, the embedding dimension to 64, the hyperedge number to 32. All experiments were conducted on a single NVIDIA A100 GPU. We evaluate model performance using the standard HR and NDCG metrics from explainable Neuro-Symbolic recommendation\cite{FENCR}. All reported results are the average of five independent runs. 

\section{Probabilistic Formulation of Front-door Adjustment}
This section provides the formal probabilistic parameterization of the two-step Front-door Adjustment (FDA) implemented in our CNRE model. Directly applying the theoretical FDA formula in recommender systems presents two major challenges. First, it requires summing over all possible mediators, which is computationally intractable in the high-dimensional, massive spaces of recommender systems \cite{dccf}. Second, the theoretical do-operator requires an idealized intervention, whereas our model can only perform computation on biased, observational data \cite{causalID}.

To address these challenges, CNRE pioneers a logic operation-based approximation paradigm. It treats the neuro-symbolic operations on endogenous logic from user behavior chains as a principled simulation of the do-operator, thereby generating the causal mediator $M$ that approximates an ideal state isolated from confounding effects. Specifically, we detail how this approximation parameterizes the key conditional probabilities, linking the abstract causal theory to our neural computation modules.

\subsection{Parameterizing $P(M|do(A,T))$}

The first step of FDA, estimating the posterior of the mediator $M$ given an intervention on the behavior chain $(A,T)$, is governed by a hierarchical decision mechanism in CNRE. A Dispatcher function first analyzes the composition of the behavior chain to determine its preference strength, $s \in$\{strong, medium, weak\}. The parameterization is then conditioned on this strength $s$, with a path-specific neuro-logic operator generating a deterministic causal mediator embedding $e_m^b$. This process is designed to find the most probable state of the mediator's posterior distribution. The dispatcher identifies the most likely reasoning mode based on the behavior chain, and the subsequent deterministic function provides a tractable estimate of this mode:

Case 1: Strong Preference ($s$=strong): When the behavior chain is complete, the mediator $e_m^b$ is generated deterministically by a function $f_{strong}$, which concatenates the final calibrated user and item embeddings $e_u^{b_t}, e_i^{b_t}$. This output represents a point estimate of the posterior:

\begin{equation}
e_m^b = f_{strong}(e_u^{b_t} \oplus e_i^{b_t})
\end{equation}

Case 2: Medium Preference ($s$=medium): If the purchase confidence is below a threshold $\tau$, the confirmatory conjunction module, $f_{medium}$ (an MLP as defined in Eq.(9)), is initiated. This operator retrieves a collaborative embedding $S_{col}^b$ via $\text{ANNS}$ from the collaborative embedding space. The final mediator $e_m^{b_t}$ is a deterministic function of the user embedding, the target item's collaborative embedding, and the retrieved collaborative embedding:

\begin{equation}
e_m^b = f_{medium}(e_u^b \oplus e_{i,col}^b \oplus S_{col}^b)
\end{equation}

Case 3: Weak Preference ($s$=weak): The supplementary disjunction module, $f_{weak}$ (an MLP as defined in Eq.(11)), is initiated. Critically, this operator retrieves a semantic embedding $S_{sem}^b$ by performing $\text{ANNS}$ in the hypergraph semantic embedding space. The final mediator $e_m^b$ is then a deterministic function of the user embedding, the target item's semantic embedding, and the retrieved semantic embedding:

\begin{equation}
e_m^b= f_{weak}(e_u^b \oplus e_{i,sem}^b \oplus S_{sem}^b)
\end{equation}

\subsection{Parameterizing $P(Y|do(M))$}
The second step of FDA, estimating the probability of the outcome $Y$ given an intervention on the mediator $M$, is directly modeled by our Prediction Module. As this module's input is solely the causal mediator $e_m^b$,  which serves as our estimate of the mediator's most likely state, it directly estimates the causal effect of this specific mediator state on $Y$. We parameterize the probability of a positive recommendation ($Y$=1) as:

\begin{equation}
P(Y=1 | do(M=e_m^b)) \approx \sigma(\text{MLP}(e_m^b))
\end{equation}

In summary, CNRE's parameterization is a principled, two-stage deterministic approximation of the intractable FDA formula. Instead of statistically modeling the full mediator distribution, our paradigm uses the endogenous logic of the behavior chain to first identify the most probable reasoning path, effectively estimating the mode of the mediator distribution. The deterministic output of the corresponding neural-logic operator then serves as a tractable point estimate for this mode. This principled selection process is what allows our model to remain both computationally tractable and highly explainable.

\section{Comparison of Explainability Paradigms}
To further clarify the methodological distinctiveness of CNRE within the broader research landscape, this section provides a comparison of existing recommendation paradigms, as summarized in Table 6. Existing explainable recommendations predominantly rely on external side information, which often creates a bottleneck due to data sparsity or privacy concerns. For instance, review-based methods (InVAE \cite{review1}, PEPLER \cite{review2}) require rich textual feedback, knowledge-aware models (CAFE \cite{KG1}, GrEA \cite{KG2}) depend on comprehensive knowledge graphs, and logic-rule approaches (FENCR \cite{FENCR}, NS-ICF \cite{NS-ICF}) necessitate explicit item attributes.

Even advanced paradigms face similar constraints: Causal approaches like CountER \cite{CountER} focus on identifying influential interactions without explicitly modeling the underlying reasoning logic, while others (e.g., CERS \cite{CERS}) depend on feature-level interventions on external attributes. Meanwhile, while LLM-based approaches (CausalX \cite{llm1} and LLMXRec \cite{llm2}) incur high computational costs and rely on vast pre-trained external knowledge. On the other end of the spectrum, traditional multi-behavior recommendations (CRGCN \cite{CRGCN}, S-MBRec \cite{s-mbrec}, BCIPM \cite{bcipm}) effectively capture interactions but remain entirely black-box, failing to transparently model the decision-making process. In contrast, CNRE introduces a novel paradigm by uncovering endogenous logic solely from user behavior chains. This approach enables multi-level explainability (from model design to results) without the need for any external information, ensuring both high generalizability and transparency.

\begin{table}[t] 
\centering
\small 
\setlength{\tabcolsep}{5pt} 
\caption{Comparison between CNRE and related works across several key characteristics.}
\label{tab:model_comparison}
\begin{tabular}{@{} c >{\centering\arraybackslash}p{1.3cm} >{\centering\arraybackslash}p{1.3cm} >{\centering\arraybackslash}p{1.3cm} >{\centering\arraybackslash}p{1.3cm} @{}}
\toprule
\textbf{Model} & \textbf{External Info} & \textbf{Model Design} & \textbf{Decision Process} & \textbf{Rec. Result} \\ 

\midrule
CRGCN & \ding{55} & \ding{55} & \ding{55} & \ding{55} \\
S-MBRec & \ding{55} & \ding{55} & \ding{55} & \ding{55} \\
BCIPM & \ding{55} & \ding{55} & \ding{55} & \ding{55} \\

\midrule
InVAE & \checkmark & \ding{55} & \checkmark & \ding{55} \\
PEPLER & \checkmark & \ding{55} & \checkmark & \ding{55} \\
CAFE & \checkmark & \checkmark & \checkmark & \checkmark \\
GrEA & \checkmark & \ding{55} & \checkmark & \ding{55} \\
CausalX & \checkmark & \checkmark & \checkmark & \ding{55} \\
LLMXRec & \checkmark & \checkmark & \checkmark & \ding{55} \\
\midrule
CountER & \ding{55} & \checkmark & \checkmark & \ding{55} \\
CERS & \checkmark & \checkmark & \checkmark & \ding{55} \\

\midrule
NS-ICF & \checkmark & \ding{55} & \checkmark & \checkmark \\
FENCR & \checkmark & \ding{55} & \checkmark & \checkmark \\

\midrule
\textbf{CNRE} & \textbf{\ding{55}} & \textbf{\checkmark} & \textbf{\checkmark} & \textbf{\checkmark} \\
\bottomrule
\end{tabular}
\end{table}


%

\end{document}